# MBA: Mini-Batch AUC Optimization

San Gultekin, Avishek Saha, Adwait Ratnaparkhi, and John Paisley

*Abstract*—Area under the receiver operating characteristics curve (AUC) is an important metric for a wide range of signal processing and machine learning problems, and scalable methods for optimizing AUC have recently been proposed. However, handling very large datasets remains an open challenge for this problem. This paper proposes a novel approach to AUC maximization, based on sampling mini-batches of positive/negative instance pairs and computing U-statistics to approximate a global risk minimization problem. The resulting algorithm is simple, fast, and learning-rate free. We show that the number of samples required for good performance is independent of the number of pairs available, which is a quadratic function of the positive and negative instances. Extensive experiments show the practical utility of the proposed method.

*Index Terms*—Receiver operating characteristics, area under curve, U-statistics, mini batch, convex optimization, matrix concentration inequalities.

## I. INTRODUCTION

Bipartite ranking has an important place in signal processing and machine learning. Given a set of positive and negative inputs, the bipartite ranking problem is concerned with building a scoring system, such that the positives are ranked higher than the negatives. This problem has been studied extensively in signal processing literature [1], in particular for radars, as missing the presence of a target (miss detection) could have dire consequences. In such a setting, a received signal is assigned a score, which is then compared against a threshold to decide if the target is present. The receiver operating characteristics (ROC) curve plots the ratio of true positives (detection) to false positives (false alarm) as a function of this threshold, and provides information about the behavior of the system. The area under the ROC curve (AUC) is a threshold-independent metric which measures the fraction of times a positive instance is ranked higher than a negative one. Therefore, it is a natural measure for the bipartite ranking accuracy.

The bipartite ranking problem and ROC curves are not limited to traditional applications. In fact, these tools are ubiquitous in modern machine learning problems. Important subfields include cost-sensitive learning and imbalanced data processing. In the latter, one is given a dataset with binary labels in which the ratio of positive to negative samples is very low. This means a classifier which predicts all incoming instances to be negative will have very high prediction accuracy. On the other hand, it will have an AUC of zero. This is worse than random guessing, which would give 0.5, and so the AUC in a sense may be a better choice for performance metric, and devising methods to achieve higher AUC is a meaningful goal. For this reason, the AUC metric is heavily used for website ad click prediction problems [2], where only a very small fraction of web history contains ads clicked by visitors. In this case, a system with high AUC is the one which can distinguish the ads that are "interesting" for a user from the rest, whereas a simple classifier that maximizes prediction accuracy may simply predict "not interesting."

Given that AUC is the primary performance measure for many problems, it is useful to devise algorithms that directly optimize this metric during the training phase. AUC optimization has been studied within the context of well-known machine learning methods, such as support vector machines [3], boosting [4], and decision trees [5]. However, most of these traditional approaches do not scale well as the size of the dataset grows, since AUC is defined over positive/negative *pairs*; this has a quadratic growth of $\mathcal{O}(N_+N_-)$. Moreover, the AUC itself is a sum of indicator functions, and its direct optimization results in NP-hard problems.

Recent research in this direction increasingly focuses on convex surrogate loss functions to represent the AUC. This enables one to use stochastic gradient methods to efficiently learn a ranking function [6]. The first work in this direction is [7], where an online AUC maximization method based on proxy hinge loss is proposed. Later, [8] use the pairwise squared loss function, which eliminates the need for buffering previous instances; [9] propose adaptive gradient/subgradient methods which can also handle sparse inputs, while [10], [11] consider the nonlinear AUC maximization problem using kernel and multiple-kernel methods. Most recently, [12] focuses on scalable kernel methods.

While these approaches can significantly increase scalability, for very large datasets their sequential nature can still be problematic. One widely used technique -particularly for training deep neural networks on large datasets [13]- is processing data in mini-batches. AUC maximization method which can utilize mini-batch processing is thus desirable. In this paper we propose a novel algorithm for fast AUC optimization. Our approach, called Mini-Batch AUC Optimization (MBA) is based on a convex relaxation of the AUC function. However instead of using stochastic gradients, it uses first and second order U-statistics of pairwise differences. A distinctive feature of our approach is it being *learning-rate free*, contrary to mini-batch gradient descent methods. This is important, as tuning the step size *a priori* is a difficult task, and generic approaches such as cross-validation are inefficient, when the dataset is large.

One of the main challenges of AUC optimization is, even if convex relaxation is applied the resulting problem is defined over pairs of positive/negative samples, and the optimization has a sample cost of $\mathcal{O}(N_+N_-)$. This grows prohibitively large even for moderate datasets. Since mini-batch optimization is based on sub-sampling it is important to understand the behavior of MBA as a function of sample size. Our theoretical analysis reveals that, the solution returned by MBA concentrates around the batch solution that utilizes the *entire pair ensemble* provided by the data, with an exponential tail bound. Unlike previous work, our proofs are based on the more recent results on matrix concentration [14]; and they are quite straightforward. In terms of the related work, while U-processes for ranking problems have previously been explored by [15]. scalable mini-batch algorithms using U-statistics have not been developed. The nearest work of [12] uses mini-batch techniques, but for gradient descent.

We organize the paper as follows. In Section 2 we review the binary hypothesis testing problem, which provides a generative model framework for the AUC optimization problem. Here we discuss the learning approach and contrast it with the detection-theoretic framework, and in particular discuss how a linear ranker can be regarded as a high signal-to-noise ratio (SNR) approximation. We develop MBA in Section 3 and show theoretical results in Section 4. Section 5 contains extensive experiments including a simulation study, fifteen datasets from UCI/LIBSVM repositories, and three large scale web click data. We conclude in Section 6.

## II. BACKGROUND: TWO FRAMEWORKS

In this section we review the ROC curves and AUC optimization through the lens of two different frameworks. Firstly, the signal detection framework is concerned with a probabilistic setup, where the optimal solution with maximum AUC can be obtained in analytical form. In contrast, the statistical learning framework assumes that the probability distributions generating the observations are unknown to the modeler. Here we discuss the use of empirical AUC loss, along with the specific case on linear ranking functions which can be viewed as a high signal-to-noise ratio (SNR) approximation of the signal detection setup. We finally discuss the convex relaxation approach to AUC optimization.

### A. Detection Theoretic Framework

A widely studied problem in signal detection is the binary hypothesis testing. Here, a received signal $\boldsymbol{X}$ is assumed to have distribution under two different hypotheses as

$$\mathcal{H}^+ : \boldsymbol{X} \sim \mathcal{P}^+ \quad , \quad \mathcal{H}^- : \boldsymbol{X} \sim \mathcal{P}^- \qquad (1)$$

This setting arises frequently in many different applications, such as radar systems and communication channels [1]. In this setup, it is commonly assumed that the generating distributions $\mathcal{P}^+$ and $\mathcal{P}^-$ are known. Then, the aim is to design a decision rule, or *detector* $\Gamma(\boldsymbol{X})$ which minimizes a given error metric. For example, when the prior probabilities and the cost of making a decision is known, a Bayes detector is the optimal choice. On the other hand, a minimax rule can be applied to minimize the worst case error when the prior distributions are unknown. For our purposes the most interesting case is that of *Neyman-Pearson (NP)* hypothesis testing, where neither the priors are known, nor the costs are directly available. In this case the optimal detector is designed based on the following two metrics: detection and false alarm. For a given detector $\Gamma(\boldsymbol{X})$ they are defined as

$$\text{Detection: } P_D(\Gamma) = \int \Gamma(\boldsymbol{x})\, p^+(\boldsymbol{x})\, d\boldsymbol{x}$$

$$\text{False Alarm: } P_F(\Gamma) = \int \Gamma(\boldsymbol{x})\, p^-(\boldsymbol{x})\, d\boldsymbol{x} \qquad (2)$$

Two immediate observations follow: (i) The metrics measure the performance of the detector itself, so they are a function of $\Gamma$. (ii) The detector $\Gamma(\boldsymbol{x})$, in turn, is a mapping from the observed signal $\boldsymbol{x}$ to the hypotheses $\mathcal{H}^+/\mathcal{H}^-$ encoded as 1/0. As the names imply, detection is the probability of correctly choosing the positive hypothesis, whereas false alarm is choosing the positive whereas the correct hypothesis was the negative. In general, the positive hypothesis corresponds to the presence of a target/message, whereas the negative one indicates absence, hence the names.

In NP hypothesis testing, the optimal detector is the solution to the optimization

$$\Gamma'(\boldsymbol{x}) := \arg\max_{\Gamma} P_D(\Gamma) \quad s.t. \quad P_F(\Gamma) \leq \alpha \ . \qquad (3)$$

We therefore seek the detector with highest detection probability while setting a limit on the false alarm rate ($0 \leq \alpha \leq 1$). Note that, without this limit (i.e. $\alpha = 1$) we can use a trivial decision rule that maps all observations to positive hypothesis and obtain $P_D(\Gamma) = 1$. The solution of this optimization is given by the following.

**Lemma 1 (Neyman-Pearson, [1]):** For $\alpha$, let $\Gamma$ be any decision rule with $P_D(\Gamma) \leq \alpha$ and let $\Gamma'$ be the decision rule of form

$$\Gamma'(\boldsymbol{x}) = \begin{cases} 1 & \text{if } p_1(\boldsymbol{x}) > \eta\, p_0(\boldsymbol{x}) \\ \gamma(\boldsymbol{x}) & \text{if } p_1(\boldsymbol{x}) = \eta\, p_0(\boldsymbol{x}) \\ 0 & \text{if } p_1(\boldsymbol{x}) < \eta\, p_0(\boldsymbol{x}) \end{cases} \qquad (4)$$

where $\eta \geq 0$ and $0 \leq \gamma(\boldsymbol{x}) \leq 1$ are chosen such that $P_F(\Gamma') = \alpha$. Then $P_D(\Gamma') \geq P_D(\Gamma)$.

We note that a detector that is optimal in the NP sense satisfies the false alarm inequality on the boundary. The structure in Eq. (4) reveals that, for any given input $\boldsymbol{x}$ the detector computes a *score* based on the likelihood ratio $p_1(x)/p_0(x)$ and compares it to a threshold. Since the ROC curve is the plot of detection vs false alarm, when $\mathcal{P}^+$ and $\mathcal{P}^-$ are known, the likelihood ratio function can be used to obtain scores with maximum AUC, whose functional form is given by

$$\text{AUC} = \mathbb{E}_{\substack{\boldsymbol{x}^+ \sim \mathcal{P}^+ \\ \boldsymbol{x}^- \sim \mathcal{P}^-}} \Big[\, \mathbb{1}\{f(\boldsymbol{x}^+) - f(\boldsymbol{x}^-) > 0\} \Big]. \qquad (5)$$

This expectation is the probability that a positive instance is ranked higher than negative instance.[1] The framework outlined in this section is displayed in Figure 1, left panel. In general, the likelihood ratio test would yield non-linear decision boundaries. In the next section we introduce the statistical learning approach, and discuss the significance of linear boundaries.

---

[1] When positive and negative instances overlap in the input space, ties are set to $1/2$.

## B. Convex Relaxation for AUC

The Neyman-Pearson lemma shows how one can maximize the AUC using likelihood ratio scoring, even if the AUC is a sum of indicator functions and its direct optimization is NP-hard. However for the subsequent development it will be necessary to apply convex relaxation to the empirical AUC loss (defined in next section), in order to get a tractable optimization problem. We therefore discuss the convex surrogate loss functions next.

Replacing $\mathbb{1}[f(\boldsymbol{x}^+) - f(\boldsymbol{x}^-) > 0]$ in Eq. (10) with the pairwise convex surrogate loss $\phi(\boldsymbol{x}^+, \boldsymbol{x}^-) = \phi(f(\boldsymbol{x}^+) - f(\boldsymbol{x}^-))$, the aim is now to minimize the $\phi$-risk [16]

$$R_\phi(f) = \mathbb{E}_{\substack{\boldsymbol{x}^+ \sim \mathcal{P}^+ \\ \boldsymbol{x}^- \sim \mathcal{P}^-}} \left[ \phi(f(\boldsymbol{x}^+) - f(\boldsymbol{x}^-)) \right]. \qquad (6)$$

This is the Bayes risk of the scoring function [17]. There are many possible choices for surrogate function; some common choices are the pairwise squared loss (PSL), pairwise hinge loss (PHL), pairwise exponential loss (PEL), and pairwise logistic loss (PLL) [18]:

$$\phi_{\text{PSL}}(t) = (1-t)^2, \quad \phi_{\text{PHL}}(t) = \max(0, 1-t), \qquad (7)$$
$$\phi_{\text{PEL}}(t) = \exp(-t), \quad \phi_{\text{PLL}}(t) = \log(1 + \exp(-t)),$$

where $t := f(\boldsymbol{x}_i^+) - f(\boldsymbol{x}_j^-)$ is the pairwise scoring difference. Among the recent works on AUC optimization, [7] and [19] use PHL, whereas [8] and [9] focus on PSL. On the other hand, all these studies are focused on deriving a stochastic-gradient based algorithm. In this paper, we use the PSL function for two reasons: (i) its consistency with the original AUC loss has been shown by [18], and (ii) the structure of PSL allows for a mini-batch algorithm, for which theoretical guarantees can be derived. Unlike stochastic-gradient methods, this formulation is learning-rate free, which quite notably increases its practicality.

We now take one further step and assume that the scoring function is linear in the original input space; however we will discuss nonlinear extensions in Section IV. In this case we have the further simplification $f(\boldsymbol{x}) = \boldsymbol{w}^\top \boldsymbol{x}$ and the $\phi$-risk becomes

$$\begin{aligned} R_\phi(f) &= \mathbb{E}_{\substack{\boldsymbol{x}^+ \sim \mathcal{P}^+ \\ \boldsymbol{x}^- \sim \mathcal{P}^-}} \left[ \left( 1 - \boldsymbol{w}^\top (\boldsymbol{x}_i^+ - \boldsymbol{x}_j^-) \right)^2 \right] \\ &= 1 - 2\boldsymbol{w}^\top \mathbb{E}\left[ (\boldsymbol{x}_i^+ - \boldsymbol{x}_j^-) \right] \\ &\quad + \boldsymbol{w}^\top \mathbb{E}\left[ (\boldsymbol{x}_i^+ - \boldsymbol{x}_j^-)(\boldsymbol{x}_i^+ - \boldsymbol{x}_j^-)^\top \right] \boldsymbol{w} \\ &= 1 - 2\boldsymbol{w}^\top \boldsymbol{\mu} + \boldsymbol{w}^\top \boldsymbol{\Sigma} \boldsymbol{w}, \end{aligned} \qquad (8)$$

where we define $\boldsymbol{x}_{ij} := (\boldsymbol{x}_i^+ - \boldsymbol{x}_j^-)$, and $\boldsymbol{\mu} = \mathbb{E}[\boldsymbol{x}_{ij}]$ and $\boldsymbol{\Sigma} = \mathbb{E}[\boldsymbol{x}_{ij} \boldsymbol{x}_{ij}^\top]$ are the first and second moments of $\boldsymbol{x}_{ij}$. We finally define the solution to the $\phi$-risk minimization problem as

$$\boldsymbol{w}^\star = \arg\min_{\boldsymbol{w}} \frac{1}{2} \boldsymbol{w}^\top \boldsymbol{\Sigma} \boldsymbol{w} - \boldsymbol{w}^\top \boldsymbol{\mu}, \qquad (9)$$

where we multiply by 1/2 for notation reasons. Note that $\boldsymbol{\mu}$ and $\boldsymbol{\Sigma}$ characterize the first and second order statistics of the pairwise differences. This is important because the quality of bipartite ranking does not directly depend on the positive and negative features, but the differences between them. This observation forms the basis of our mini-batch algorithm. Finally, by definition $\boldsymbol{\Sigma}$ is positive semi-definite and when it is positive definite, there is a unique $\boldsymbol{w}^\star$ that satisfies Eq. (9).

## C. Statistical Learning Framework

For the setup described by Eq. (1) assume that the distributions $\mathcal{P}^+ / \mathcal{P}^-$ generating the features are unknown, but instead we are given samples generated from them, organized into a training set $\mathcal{X}$ along with the labels $\mathcal{Y} = \{\pm 1\}$. The task is once again, to design a score function which should yield high AUC. A score function in its most general sense is a mapping $f : \mathcal{X} \to \mathbb{R}$ from a given input to a continuous real number, which induces a total order on the data points. [2] In the learning setting, the AUC will be measured on a separately provided test set, which shows the *generalization ability*. The ideal solution is once again the NP detector; however in this case it cannot be computed as the generating distributions are unknown. Following the standard statistical learning approach, we can substitute the *empirical AUC* as the objective function, which we seek to maximize over the training set

$$\text{AUC} = \sum_{i=1}^{N_+} \sum_{j=1}^{N_-} \mathbb{1}\{f(\boldsymbol{x}_i^+) - f(\boldsymbol{x}_j^-) > 0\}. \qquad (10)$$

As mentioned before, direct optimization of this metric gives rise to an NP-hard problem since the objective in Eq. (10) is a sum of indicator functions; furthermore the number of pairs grow quadratically with the training data. To sidestep this difficulty, a surrogate loss function $\phi(\cdot)$ can be chosen to replace the Eq. (10), as we discuss in the previous section. In particular, the empirical $\phi$-risk (c.f. Eq. (8)) is

$$\widehat{R}_\phi(f) = \frac{1}{2N_+ N_-} \sum_{i=1}^{N_+} \sum_{j=1}^{N_-} \phi(f(\boldsymbol{x}_i^+) - f(\boldsymbol{x}_j^-)), \qquad (11)$$

and this will be the optimization objective of our proposed MBA, introduced in the next section.

Finally, the right panel of Figure 1 displays the statistical learning approach to AUC optimization, where the $\phi$-risk is minimized under the linear classifier assumption. In contrast to the signal detection problem of left panel, here we only have access to samples, instead of the class-conditional densities. As the figure suggests, the linear scoring function assumption is meaningful when the two classes are linearly separable. This, in turn, depends on the separation between the two class-conditional distributions, which is measured by signal-to-noise ratio (SNR) in signal processing. Therefore, applying a linear machine learning model can be regarded as assuming sufficiently high signal-to-noise ratio in the given dataset.

## III. MINI-BATCH AUC OPTIMIZATION

Using the pairwise squared loss function we obtain a convex optimization problem in place of the original NP-hard problem. However, Eq. (9) is still difficult since computing the first and second order statistics rely on the knowledge of the data generating distribution $\mathcal{P}$. In practical settings, we are only given a set of positive and negative instances sampled from $\mathcal{P}$, written $\mathcal{S}^+ = \{\boldsymbol{x}_1^+, \ldots, \boldsymbol{x}_{N_+}^+\}$, $\mathcal{S}^- = \{\boldsymbol{x}_1^-, \ldots, \boldsymbol{x}_{N_-}^-\}$. We therefore substitute the empirical risk in Eq. (11) which can be more easily optimized. The term $N := N_+ N_-$ corresponds to the total number of pairs in the data. Similar to the $\phi$-risk, optimizing the empirical risk yields a convex problem, but the number of pairs grows quadratic in the number of data points. Therefore, even for moderate datasets, minimizing the empirical risk in Eq. (11) becomes intractable.

---

[2] The likelihood ratio in NP detector is one example of such scoring function.

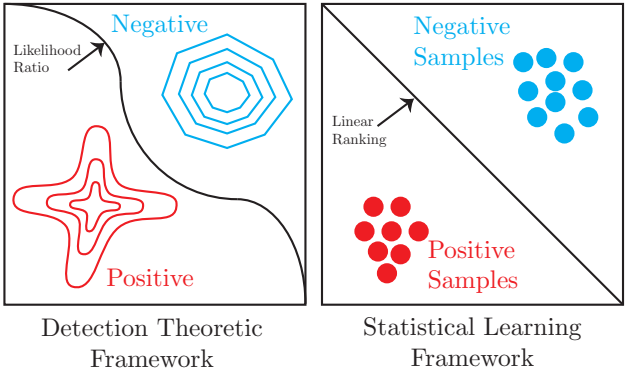

Fig. 1. A cartoon illustration of the signal detection (left) and statistical learning (right) frameworks for AUC maximization.

We substitute the PSL and functional form of linear classifier into the empirical risk to obtain

$$\widehat{R}_\phi(\boldsymbol{w}) = -\boldsymbol{w}^\top \left[ \frac{1}{N_+ N_-} \sum_{i=1}^{N_+} \sum_{j=1}^{N_-} (\boldsymbol{x}_i^+ - \boldsymbol{x}_j^-) \right] \quad (12)$$
$$+ \frac{1}{2} \boldsymbol{w}^\top \left[ \frac{1}{N_+ N_-} \sum_{i=1}^{N_+} \sum_{j=1}^{N_-} (\boldsymbol{x}_i^+ - \boldsymbol{x}_j^-)(\boldsymbol{x}_i^+ - \boldsymbol{x}_j^-)^\top \right] \boldsymbol{w}$$

and define

$$\boldsymbol{\mu}_N = \frac{1}{N_+ N_-} \sum_{i=1}^{N_+} \sum_{j=1}^{N_-} (\boldsymbol{x}_i^+ - \boldsymbol{x}_j^-)$$
$$\boldsymbol{\Sigma}_N = \frac{1}{N_+ N_-} \sum_{i=1}^{N_+} \sum_{j=1}^{N_-} (\boldsymbol{x}_i^+ - \boldsymbol{x}_j^-)(\boldsymbol{x}_i^+ - \boldsymbol{x}_j^-)^\top. \quad (13)$$

The variables in Eq. (13) are sample approximations to the first and second moments of the pairwise differences, which are substituted for $\boldsymbol{\mu}$ and $\boldsymbol{\Sigma}$ in Eq. (9). Overall, the optimization problem to be solved is

$$\boldsymbol{w}^\star = \arg\min_{\boldsymbol{w}} \frac{1}{2} \boldsymbol{w}^\top \boldsymbol{\Sigma}_N \boldsymbol{w} - \boldsymbol{w}^\top \boldsymbol{\mu}_N + R_{\text{EL}}(\boldsymbol{w}), \quad (14)$$

where $R_{\text{EL}}(\boldsymbol{w}) := \lambda_1 \|\boldsymbol{w}\|_1 + (1/2)\lambda_2 \|\boldsymbol{w}\|_2^2$ is the elastic net regularizer [20], which we add to prevent overfitting. Note that, unlike Eq. (14), there is a unique optimum since the elastic net penalty makes the objective strictly convex. In addition, this regularizer encourages solution which combines small $\ell_2$ norm with sparsity. By substituting appropriate values for $\lambda_1$ and $\lambda_2$ we also recover ridge and lasso regression. In this paper, we report results for all three cases.

Since it is impractical to use all $N$ samples, we propose to use mini-batches to obtain estimates of the moments. This is a simple process which only requires the computation of U-statistics. Note that, given a parameter $\theta$ and symmetric measurable function $h$ which satisfies $\theta = h(X_1, \ldots, X_m)$, the corresponding U-statistic is given by

$$U_n = \binom{n}{m}^{-1} \sum_{C_{n,m}} h(X_1, \ldots, X_n), \quad (15)$$

where $C_{n,m}$ is the set of all length-$m$ combinations with increasing indices. As the name implies the U-statistics are unbiased, so $\theta = \mathbb{E}[U_n]$, and provide best unbiased estimators [21]. On the other hand, a U-statistic of the second moment matrix $\boldsymbol{\Sigma}_N$ also provides a building block to get an exponential concentration bound [14]. Our theoretical analysis will use this property.

### A. The MBA algorithm

We now describe the proposed MBA algorithm. Let $T$ be the total number of rounds. At round $t$ we sample $B$ positive and $B$ negative samples from the entire population with replacement. Let $\mathcal{S}_t^+$ and $\mathcal{S}_t^-$ be the arrays of sample indices and let $\mathcal{S}_t$ be the array of pairs stored as tuples of the form $(\mathcal{S}_t^+(i), \mathcal{S}_t^-(i))$—note that we do not form the Cartesian product. The expressions for U-statistics of the first and second moments simplify from Eq. (15) as

$$\boldsymbol{\mu}_t := \frac{1}{B} \sum_{(i,j) \in \mathcal{S}_t} (\boldsymbol{x}_i^+ - \boldsymbol{x}_j^-)$$
$$\boldsymbol{\Sigma}_t := \frac{1}{B} \sum_{(i,j) \in \mathcal{S}_t} (\boldsymbol{x}_i^+ - \boldsymbol{x}_j^-)(\boldsymbol{x}_i^+ - \boldsymbol{x}_j^-)^\top. \quad (16)$$

Finally let $S = BT$ denote the total number of pairs sampled by our algorithm. We also introduce the notation $\mathcal{S}_{1:T}$ for the entire array of pairs sampled during all rounds. The overall moment approximations are therefore

$$\boldsymbol{\mu}_S := \frac{1}{BT} \sum_{(i,j) \in \mathcal{S}_{1:T}} (\boldsymbol{x}_i^+ - \boldsymbol{x}_j^-)$$
$$\boldsymbol{\Sigma}_S := \frac{1}{BT} \sum_{(i,j) \in \mathcal{S}_{1:T}} (\boldsymbol{x}_i^+ - \boldsymbol{x}_j^-)(\boldsymbol{x}_i^+ - \boldsymbol{x}_j^-)^\top, \quad (17)$$

and the optimization problem constructed by MBA is

$$\boldsymbol{w}_S^\star = \arg\min_{\boldsymbol{w}} \frac{1}{2} \boldsymbol{w}^\top \boldsymbol{\Sigma}_S \boldsymbol{w} - \boldsymbol{w}^\top \boldsymbol{\mu}_S + R_{\text{EL}}(\boldsymbol{w}). \quad (18)$$

This is the function MBA aims to construct and solve, which itself is an approximation to the global risk minimization problem in Eq. (9). On the other hand, stochastic gradient-based approaches make local gradient approximations to the global function and seek a solution that way. As we will show in the experiments, this is an important difference and MBA can find better solutions since it constructs a global problem first. We summarize the proposed MBA in Algorithm 1.

Mini-batch optimizations are heavily employed in machine learning, including training of deep neural networks [13] and scalable Bayesian inference [22]. The main benefit of using mini-batches is it is significantly faster compared to the sequential approach. Online methods for optimizing AUC, however, require a sequential processing, as the parameters are updated per input. This is the main reason MBA offers a significant improvement in speed. In addition to this MBA

**Algorithm 1** Mini-Batch AUC Optimization (MBA)

1: **Require:** $B$, $T$, $\lambda_1$, $\lambda_2$
2: **Input:** $\boldsymbol{X}^+$, $\boldsymbol{X}^-$
3: **Output:** $\boldsymbol{w}^\star$
4: Initialize $\boldsymbol{\mu}_S = \boldsymbol{0}$ and $\boldsymbol{\Sigma}_S = \boldsymbol{0}$.
5: **for** $t = 1, \ldots, T$ **do**
6:    Construct index set $\mathcal{S}_t^+$ of size $B$ sampling positive examples uniformly with replacement.
7:    Construct index set $\mathcal{S}_t^-$ of size $B$ sampling negative examples uniformly with replacement.
8:    Construct $\mathcal{S}_t(i) = (\mathcal{S}_t^+(i), \mathcal{S}_t^-(i))$, $i = 1, \ldots, B$.
9:    $\boldsymbol{\mu}_S \leftarrow \boldsymbol{\mu}_S + \frac{1}{BT} \sum_{(i,j) \in \mathcal{S}_t} (\boldsymbol{x}_i^+ - \boldsymbol{x}_j^-)$
10:   $\boldsymbol{\Sigma}_S \leftarrow \boldsymbol{\Sigma}_S + \frac{1}{BT} \sum_{(i,j) \in \mathcal{S}_t} (\boldsymbol{x}_i^+ - \boldsymbol{x}_j^-)(\boldsymbol{x}_i^+ - \boldsymbol{x}_j^-)^\top$
11: **end for**
12: $\boldsymbol{w}^\star = \arg\min_{\boldsymbol{w}} \frac{1}{2} \boldsymbol{w}^\top \boldsymbol{\Sigma}_S \boldsymbol{w}^\top - \boldsymbol{w}^\top \boldsymbol{\mu}_S + \lambda_1 \|\boldsymbol{w}\|_1 + \frac{\lambda_2}{2} \|\boldsymbol{w}\|_2^2$

offers several other advantages. Since sampling pairs and computing U-statistics is an isolated process, MBA can easily be distributed across machines, which can work in an asynchronous manner. Therefore MBA is suitable for cluster computing. Secondly, streaming and/or nonstationary data processing can be incorporated into the MBA framework, as it can process streams as blocks and give larger weights to more recent ones.

## IV. THEORETICAL ANALYSIS

Solving the regularized empirical risk minimization problem in Eq. (14) requires processing $N$ pairwise samples. As this number grows quadratically with the number of positive and negative samples, it is often not possible to do this exactly. The proposed MBA addresses this problem by approximating the $N$-pair problem with an $S$-pair one, where $S$ samples are collected in mini-batches and the total number of processed samples is much less than $N$. This results in the problem in Eq. (18). Clearly the success of this approach depends on how well the second problem approximates the first. In this section we derive performance guarantees.

First, define

$$\mathcal{S}^\star = \arg\min_{\boldsymbol{w}} \tfrac{1}{2} \boldsymbol{w}^\top \boldsymbol{\Sigma} \boldsymbol{w} - \boldsymbol{w}^\top \boldsymbol{\mu},$$
$$\mathcal{S}_S^\star = \arg\min_{\boldsymbol{w}} \tfrac{1}{2} \boldsymbol{w}^\top \boldsymbol{\Sigma}_S \boldsymbol{w} - \boldsymbol{w}^\top \boldsymbol{\mu}_S. \quad (19)$$

$\mathcal{S}^\star$ is the set of solutions to the original empirical risk minimization problem of Eq. (9), and $\mathcal{S}_S^\star$ is the set of solutions to the optimization problem constructed by MBA as in Eq. (18), but without regularization. Also, in both cases the solution set has a single element if $\boldsymbol{\Sigma}$ and $\boldsymbol{\Sigma}_S$ are positive definite.

**Proposition 2:** If MBA is given an infinite number of pairs sampled i.i.d. from the unknown data distribution $\mathcal{P}$, then $\mathcal{S}^\star = \mathcal{S}_S^\star$ with probability 1.

**Proof:** The U-statistics of the first and second order moments defined in Eq. (16) converge almost surely due to the strong law of large numbers, i.e. $\boldsymbol{\mu}_S \overset{a.s.}{\to} \boldsymbol{\mu}$ and $\boldsymbol{\Sigma}_S \overset{a.s.}{\to} \boldsymbol{\Sigma}$. Since the function

$$f(\boldsymbol{w}; \boldsymbol{\mu}, \boldsymbol{\Sigma}) = \boldsymbol{w}^\top \boldsymbol{\Sigma} \boldsymbol{w} - 2\boldsymbol{w}^\top \boldsymbol{\mu}$$

is continuous, for any given $\boldsymbol{w}$, $P(f(\boldsymbol{w}; \boldsymbol{\mu}, \boldsymbol{\Sigma}) = f(\boldsymbol{w}; \boldsymbol{\mu}_S, \boldsymbol{\Sigma}_S)) = 1$ as a result of continuous mapping theorem [21]. Then

$$P(\mathcal{S}^\star = \mathcal{S}_S^\star) = P[\forall \boldsymbol{w} : f(\boldsymbol{w}; \boldsymbol{\mu}, \boldsymbol{\Sigma}) = f(\boldsymbol{w}; \boldsymbol{\mu}_S, \boldsymbol{\Sigma}_S)] = 1$$

follows. ∎

Therefore the loss function of MBA is consistent with the Bayes $\phi$-risk. In practice, however, it is desirable to have good performance with a sample size that is significantly smaller than $N$. We next provide a bound that the solution with $S$ samples is close to the truth with $S \ll N$. Using the difference between final costs $|\mathcal{L}_N(\boldsymbol{w}_N) - \mathcal{L}_N(\boldsymbol{w}_S)|$ results in a regret bound similar to the ones used in comparing online/batch versions of algorithms [23], whereas the Euclidean distance $d(\boldsymbol{w}_N - \boldsymbol{w}_S) = \|\boldsymbol{w}_N - \boldsymbol{w}_S\|_2^2$ measures how similar the two solutions are, and is used by recent work on matrix sketching [24]. Here we show results for the Euclidean distance in Theorem 1, but both metrics are addressed in the process.

We define the following two functions for convenience,

$$\mathcal{L}_N = \frac{1}{2} \boldsymbol{w}^\top \boldsymbol{\Sigma}_N \boldsymbol{w}^\top - \boldsymbol{w}^\top \boldsymbol{\mu}_N + \lambda_1 \|\boldsymbol{w}\|_1 + \frac{\lambda_2}{2} \|\boldsymbol{w}\|_2^2$$
$$\mathcal{L}_S = \frac{1}{2} \boldsymbol{w}^\top \boldsymbol{\Sigma}_S \boldsymbol{w}^\top - \boldsymbol{w}^\top \boldsymbol{\mu}_S + \lambda_1 \|\boldsymbol{w}\|_1 + \frac{\lambda_2}{2} \|\boldsymbol{w}\|_2^2, \quad (20)$$

and let $\boldsymbol{w}_N$ and $\boldsymbol{w}_S$ denote their unique minimizers. Since $\mathcal{L}_N(\boldsymbol{w})$ is strictly convex, for a fixed $\delta$ there exists an $\epsilon$ such that

$$|\mathcal{L}_N(\boldsymbol{w}_N) - \mathcal{L}_N(\boldsymbol{w}_S)| \leq \epsilon \implies \|\boldsymbol{w}_N - \boldsymbol{w}_S\|_2 \leq \delta. \quad (21)$$

Clearly, $\epsilon$ is the infimum of $\mathcal{L}_N(\cdot)$ over the circle centered at $\boldsymbol{w}_N$ with radius $\delta$. Consequently, one can focus on bounding the objective function, and this will yield the desired bound on the solutions. We now introduce $\ell_2$-norm bounds on data and weight vectors, and for any given input we assume that $\|\boldsymbol{x}\|_2^2 \leq R_x$.[3] Next, define the upper bound on weights such that $\max\{\|\boldsymbol{w}_N\|_2^2, \|\boldsymbol{w}_S\|_2^2\} \leq R_w$. Note here that $R_w < \infty$ is guaranteed by the $\ell_2$ regularization of elastic net. Also define the following quantities for convenience,

$$\Delta(\boldsymbol{w}) = \mathcal{L}_S(\boldsymbol{w}) - \mathcal{L}_N(\boldsymbol{w}) = \frac{1}{2} \boldsymbol{w}^\top \Delta_{\boldsymbol{\Sigma}} \boldsymbol{w} + \boldsymbol{w}^\top (\boldsymbol{\mu}_N - \boldsymbol{\mu}_S)$$
$$\Delta_{\boldsymbol{\Sigma}} = \boldsymbol{\Sigma}_S - \boldsymbol{\Sigma}_N$$
$$\Delta_\sigma = (\boldsymbol{w}_N - \boldsymbol{w}_S)^\top (\boldsymbol{\mu}_N - \boldsymbol{\mu}_S). \quad (22)$$

Here it can be seen that the objective function can be bounded through $\Delta(\boldsymbol{w})$, which in turn can be bounded through $\Delta_{\boldsymbol{\Sigma}}$ and $\Delta_\sigma$. The following lemma provides two useful concentration inequalities for this purpose.

**Lemma 3:** Let $\|\Delta_{\boldsymbol{\Sigma}}\|_2$ and $|\Delta_\sigma|$ denote the spectral and $\ell_1$ norms respectively. For $\gamma > 0$ and sample size $S$,

(i) $P(\|\Delta_{\boldsymbol{\Sigma}}\|_2 > \gamma) \leq$
$$2d \exp\left\{-S \frac{\gamma^2}{4R_x \|\boldsymbol{\Sigma}_N\|_2 + (8/3)\gamma R_x}\right\}$$

(ii) $P(|\Delta_\sigma| > \gamma) \leq$
$$2 \exp\left\{-S \frac{\gamma^2}{4R_w \|\boldsymbol{\Sigma}_N\|_2 + (8/3)\gamma \sqrt{R_x R_w}}\right\}$$

---
[3]This is a mild assumption since training data is typically normalized.

**Proof:** We will use the shorthand $\Delta_{\boldsymbol{w}} = \boldsymbol{w}_N - \boldsymbol{w}_S$, $\Delta_{\boldsymbol{\mu}} = \boldsymbol{\mu}_N - \boldsymbol{\mu}_S$, and $\Delta \boldsymbol{x}_s = (\boldsymbol{x}_i^+ - \boldsymbol{x}_j^-)$ where $\mathcal{S}[s] = (i, j)$. We also recall the Bernstein inequality for a $d \times d$ symmetric, random matrix $\boldsymbol{Z} = \sum_s \boldsymbol{E}_s$ and threshold $\gamma$

$$P[\boldsymbol{Z} > \gamma] \leq 2d \exp\left(\frac{-\gamma^2/2}{\mathbb{V}(\boldsymbol{Z}) + L\gamma/3}\right) \quad (23)$$

where $\|\boldsymbol{E}_s\| \leq L$. The scalar version is recovered by setting $d = 1$.

(i) This part follows the argument for the sample covariance estimator in [14]. For the matrix we can write $\Delta_{\boldsymbol{\Sigma}} = \boldsymbol{\Sigma}_S - \boldsymbol{\Sigma}_N = \sum_{s \in \mathcal{S}} \frac{1}{S}[\Delta \boldsymbol{x}_s \Delta \boldsymbol{x}_s^\top - \boldsymbol{\Sigma}_N]$. We denote each summand by $\boldsymbol{E}_s = \frac{1}{S}[\Delta \boldsymbol{x}_s \Delta \boldsymbol{x}_s^\top - \boldsymbol{\Sigma}_N]$. It then follows from triangle inequality that

$$\|\boldsymbol{E}_s\|_2 \leq \frac{1}{S}\left[\|\Delta \boldsymbol{x}_s \Delta \boldsymbol{x}_s^\top\|_2 + \|\boldsymbol{\Sigma}_N\|_2\right] = \frac{4R_x}{S} . \quad (24)$$

As each summand is centered and iid, the variance of sum decomposes as $\mathbb{V}(\Delta_{\boldsymbol{\Sigma}}) = \|\sum_{s \in \mathcal{S}} \mathbb{E}[\boldsymbol{E}_s^2]\|$. For a single summand the second moment can be bounded as

$$\mathbb{E}[\boldsymbol{E}_s^2] = \frac{1}{S^2} \boldsymbol{E}\left[\Delta \boldsymbol{x}_s \Delta \boldsymbol{x}_s^\top - \boldsymbol{\Sigma}_N\right]^2$$
$$= \frac{1}{S^2}\left[\boldsymbol{E}\left[\|\Delta \boldsymbol{x}_s\|_2^2 \Delta \boldsymbol{x}_s \Delta \boldsymbol{x}_s^\top\right] - \boldsymbol{\Sigma}_N^2\right]$$
$$\preceq \frac{1}{S^2}\left[2R_x \boldsymbol{\Sigma}_N - \boldsymbol{\Sigma}_N^2\right]$$
$$\preceq \frac{2R_x \boldsymbol{\Sigma}_N}{S^2} , \quad (25)$$

from which the variance inequality $\mathbb{V}(\Delta_{\boldsymbol{\Sigma}}) \leq \frac{2R_x \|\boldsymbol{\Sigma}_N\|_2}{S}$ follows. Substituting Eqs. (24) and (25) into Eq. (23) yields the result.

(ii) For this part, the scalar version of Eq. (23) can be used. We have $\Delta_\sigma = (\boldsymbol{w}_N - \boldsymbol{w}_S)^\top (\boldsymbol{\mu}_N - \boldsymbol{\mu}_S) = \sum_{s \in \mathcal{S}} \frac{1}{S}\left[(\boldsymbol{w}_N - \boldsymbol{w}_S)^\top (\boldsymbol{\mu}_N - \boldsymbol{x}_s)\right]$ where we denote each *scalar* summand as $e_s = \frac{1}{S}\left[(\boldsymbol{w}_N - \boldsymbol{w}_S)^\top (\boldsymbol{\mu}_N - \boldsymbol{x}_s)\right]$. Once again it is straightforward to verify each summand is centered and iid. An $\ell_1$-norm bound can be obtained by Cauchy-Schwarz inequality

$$|e_s| = \left|\frac{1}{S}(\boldsymbol{w}_N - \boldsymbol{w}_S)^\top (\boldsymbol{\mu}_N - \boldsymbol{x}_s)\right|$$
$$\leq \frac{1}{S}\|\boldsymbol{w}_N - \boldsymbol{w}_S\|_2 \|\boldsymbol{\mu}_N - \boldsymbol{x}_s\|_2$$
$$= \frac{4\sqrt{R_x R_w}}{S} . \quad (26)$$

For the variance we once again have the decomposition $\mathbb{V}(\Delta_\sigma) = \sum_{s \in \mathcal{S}} \mathbb{E}[e_s^2]$ and for a single term we have

$$\mathbb{E}[e_s^2] = \mathbb{E}\left[\left[\frac{1}{S}(\boldsymbol{w}_N - \boldsymbol{w}_S)^\top (\boldsymbol{\mu}_N - \boldsymbol{x}_s)\right]^2\right]$$
$$= \frac{1}{S^2}\left[\mathbb{E}[\Delta_{\boldsymbol{w}}^\top \boldsymbol{x}_s \boldsymbol{x}_s^\top \Delta_{\boldsymbol{w}}] + \frac{1}{S^2}\mathbb{E}[\Delta_{\boldsymbol{w}}^\top \boldsymbol{\mu}_N \boldsymbol{\mu}_N^\top \Delta_{\boldsymbol{w}}]\right.$$
$$\left. - \frac{1}{S^2}\mathbb{E}[\Delta_{\boldsymbol{w}}^\top \boldsymbol{x}_s \boldsymbol{\mu}_N^\top \Delta_{\boldsymbol{w}}] - \frac{1}{S^2}\mathbb{E}[\Delta_{\boldsymbol{w}}^\top \boldsymbol{\mu}_N \boldsymbol{x}_s^\top \Delta_{\boldsymbol{w}}]\right]$$
$$= \frac{1}{S^2}\Delta_{\boldsymbol{w}}^\top \boldsymbol{\Sigma}_N \Delta_{\boldsymbol{w}} - \frac{1}{S^2}\Delta_{\boldsymbol{w}}^\top \boldsymbol{\mu}_N \boldsymbol{\mu}_N^\top \Delta_{\boldsymbol{w}}$$
$$\leq \frac{2R_w \|\boldsymbol{\Sigma}_N\|_2}{S^2} \quad (27)$$

where for the last line we used the upper bounds $\|\Delta_{\boldsymbol{w}}\|_2^2 \leq 2R_w$ and $\Delta_{\boldsymbol{w}}^\top \boldsymbol{\Sigma}_N \Delta_{\boldsymbol{w}} \leq \|\Delta_{\boldsymbol{w}}\|_2^2 \|\boldsymbol{\Sigma}_N\|_2^2$ and dropped the negative term. Note that the first upper bound holds by triangle inequality, and the second one follows from the maximum eigenvalue bound of a quadratic form. Plugging Eqs. (26) and (27) into Eq. (23) we get the desired result. ∎

Using Lemma 1, we can derive the following result.

**Theorem 4:** Let $\boldsymbol{w}_S^\star$ be the solution returned by MBA using $S$ samples. For $\epsilon > 0$, if

$$S \geq \max\left\{\log(4d/p)\frac{[48R_w^2 \|\boldsymbol{\Sigma}_N\|_2 + 16\epsilon R_w]R_x}{3\epsilon^2} , \right.$$
$$\left. \log(4/p)\frac{48R_w \|\boldsymbol{\Sigma}_N\|_2 + 16\epsilon\sqrt{R_x R_w}}{3\epsilon^2}\right\} \quad (28)$$

then $\|\boldsymbol{w}_N - \boldsymbol{w}_S\|_2 \leq \delta$ with probability at least $1 - p$.

**Proof:** The starting point of the proof is the equality

$$\mathcal{L}_S(\boldsymbol{w}_N) - \mathcal{L}_S(\boldsymbol{w}_S) =$$
$$\mathcal{L}_N(\boldsymbol{w}_N) + \Delta(\boldsymbol{w}_N) - \mathcal{L}_N(\boldsymbol{w}_S) - \Delta(\boldsymbol{w}_S) . \quad (29)$$

Here by construction

$$\mathcal{L}_N(\boldsymbol{w}_S) - \mathcal{L}_N(\boldsymbol{w}_N) = \mathcal{L}_N(\boldsymbol{w}_S) - \arg\min_{\boldsymbol{w}} \mathcal{L}_N(\boldsymbol{w}) \geq 0 \quad (30)$$

$$\mathcal{L}_S(\boldsymbol{w}_N) - \mathcal{L}_S(\boldsymbol{w}_S) = \mathcal{L}_S(\boldsymbol{w}_N) - \arg\min_{\boldsymbol{w}} \mathcal{L}_S(\boldsymbol{w}) \geq 0 \quad (31)$$

and we obtain

$$0 \leq \mathcal{L}_N(\boldsymbol{w}_S) - \mathcal{L}_N(\boldsymbol{w}_N) \leq \Delta(\boldsymbol{w}_N) - \Delta(\boldsymbol{w}_S) . \quad (32)$$

The left hand size of the inequality is a direct consequence of Eq. (30). For the right-hand side note that Eq. (29) can be manipulated as

$$\mathcal{L}_N(\boldsymbol{w}_S) - \mathcal{L}_N(\boldsymbol{w}_N) = \Delta(\boldsymbol{w}_N) - \Delta(\boldsymbol{w}_S)$$
$$+ [\mathcal{L}_S(\boldsymbol{w}_S) - \mathcal{L}_S(\boldsymbol{w}_N)]$$
$$\leq \Delta(\boldsymbol{w}_N) - \Delta(\boldsymbol{w}_S) \quad (33)$$

where the second line follows from $[\mathcal{L}_S(\boldsymbol{w}_S) - \mathcal{L}_S(\boldsymbol{w}_N)] < 0$ due to Eq. (31).

It is therefore sufficient to show that $\Delta(\boldsymbol{w}_N) - \Delta(\boldsymbol{w}_S) \leq \epsilon$ with high probability; the result then follows from the strict convexity argument. Further expand this bounding term as

$$\Delta(\boldsymbol{w}_N) - \Delta(\boldsymbol{w}_S) =$$
$$\frac{1}{2}\boldsymbol{w}_N^\top \Delta_{\boldsymbol{\Sigma}} \boldsymbol{w}_N - \frac{1}{2}\boldsymbol{w}_S^\top \Delta_{\boldsymbol{\Sigma}} \boldsymbol{w}_S + \boldsymbol{w}_N^\top \Delta_{\boldsymbol{\mu}} - \boldsymbol{w}_S^\top \Delta_{\boldsymbol{\mu}} , \quad (34)$$

and consider uniformly bounding the following two terms:
- Quadratic: $\left|\frac{1}{2}\boldsymbol{w}_N^\top \Delta_{\boldsymbol{\Sigma}} \boldsymbol{w}_N - \frac{1}{2}\boldsymbol{w}_S^\top \Delta_{\boldsymbol{\Sigma}} \boldsymbol{w}_S\right| < \frac{\epsilon}{2}$
- Linear: $\left|\boldsymbol{w}_S^\top \Delta_{\boldsymbol{\mu}} - \boldsymbol{w}_N^\top \Delta_{\boldsymbol{\mu}}\right| < \frac{\epsilon}{2}$

For the quadratic term we have

$$\left|\frac{1}{2}\boldsymbol{w}_N^\top \Delta_{\boldsymbol{\Sigma}} \boldsymbol{w}_N - \frac{1}{2}\boldsymbol{w}_S^\top \Delta_{\boldsymbol{\Sigma}} \boldsymbol{w}_S\right|$$
$$\leq \frac{1}{2}\left|\boldsymbol{w}_N^\top \Delta_{\boldsymbol{\Sigma}} \boldsymbol{w}_N\right| + \frac{1}{2}\left|\boldsymbol{w}_S^\top \Delta_{\boldsymbol{\Sigma}} \boldsymbol{w}_S\right|$$
$$\leq \frac{1}{2}R_w \|\Delta_{\boldsymbol{\Sigma}}\|_2 + \frac{1}{2}R_w \|\Delta_{\boldsymbol{\Sigma}}\|_2$$
$$= R_w \|\Delta_{\boldsymbol{\Sigma}}\|_2 . \quad (35)$$

Here the first line is obtained via triangle inequality, for the second line, note that for any given quadratic form the inequality $\boldsymbol{w}^\top \Delta_{\boldsymbol{\Sigma}} \boldsymbol{w} \leq \|\boldsymbol{w}\|_2^2 \|\Delta_{\boldsymbol{\Sigma}}\|_2$ holds, as for any unit norm input $\boldsymbol{u}$, $\boldsymbol{u}^\top \Delta_{\boldsymbol{\Sigma}} \boldsymbol{u}$ is maximized at the largest eigenvalue of $\Delta_{\boldsymbol{\Sigma}}$. Equivalently, we want $\|\Delta_{\boldsymbol{\Sigma}}\|_2 \leq \epsilon/(2R_w)$ with high probability. Now applying Lemma 2-i with threshold $\gamma = \epsilon/(2R_w)$ and probability level $p/2$, we obtain the first term in Eq. (28).

For the linear term, note that this is already equal to $\Delta_\sigma$ by definition, i.e. we want to achieve $|\Delta_\sigma| \leq \epsilon/2$ with probability at least $1 - p/2$. Applying Lemma 2-ii with threshold $\gamma = \epsilon/2$ and probability level $p/2$, we obtain the second term in Eq. (28). Since both terms are bounded with probability at least $1 - p/2$, the theorem now follows from the union bound. ∎

Theorem 4 shows that the number of samples $S$ required to guarantee $\|\boldsymbol{w}_N - \boldsymbol{w}_S\|_2 < \delta$ with high probability does not depend on the total number of pairs $N = N_+ N_-$ provided. Instead the sample size grows logarithmically with the feature size. This result is useful in that, even though the total number of pairs in the data is too large, randomly sampling a small fraction guarantees a solution that is close to the true solution.

While the high SNR assumption is reasonable, in practice it is possible to have datasets that are not linearly separable; in such cases one is typically concerned with devising a nonlinear feature transform, to obtain separability. In fact, for finite-dimensional transforms Theorem 4 readily extends. Such transforms include, for example, polynomial features and conjunctions. In addition, [25] show that finite dimensional features can also be used to efficiently represent infinite dimensional kernel transforms. In more abstract terms, all these transformations are mappings from $d$ dimensions to $F$ dimensions. Given such fixed transformation, the result of Theorem 4 still holds, where we replace $d$ by $F$. Therefore MBA is equally effective where the input space is not suitable for linear ranking and we first transform the space and apply the algorithm.

## V. Experiments

In this section we conduct three types of experiments to demonstrate the performance benefits of MBA. In the first part we use simulation data from Gaussian mixtures to investigate the interplay between the signal detection and statistical learning frameworks. Here we show that MBA can achieve better performance with low number of samples, corroborating the theoretical analysis. For the second part, we experiment on 15 frequently used benchmark datasets from the UCI[4] and LIBSVM[5] repositories; these datasets cover a wide range of application domains and show MBA performs better than the competing methods overall. Finally we consider large scale click through rate (CTR) prediction problem with two publicly available commercial-size datasets with tens of millions of samples.

For comparison we use the following algorithms: MBA-$\ell_2$, MBA-$\ell_1$, MBA-EL, which represent the three variants of our mini-batch AUC optimization using ridge regression, lasso, and elastic net respectively. OLR is simply the online logistic regression and SOLR is the sparse regression algorithm presented in [2]. OAM is the first proposed online AUC maximization algorithm using stochastic gradients [7] which uses PHL. On the other hand, AdaAUC is the adaptive gradient AUC maximization algorithm in [9]. This algorithm is proposed as an improvement to one pass AUC optimization algorithm in [8]. It is shown that AdaAUC has better performance empirically. Therefore we use this version in our comparisons. Both algorithms are based on PSL. In addition to these, we implement two mini-batch stochastic gradient algorithms for large scale CTR prediction problems: MB-PHL is a mini-batch gradient descent algorithm which uses PHL. A variant of this approach is also proposed in the recent work of [12]. MB-PSL is another mini-batch gradient method that uses PSL. MB-PHL and MB-PSL can be thought of a substitutions for OAM and AdaAUC for larger datasets. This is necessary as for large number of inputs sequential processing becomes inefficient. All implementations are done in Python with NumPy and Scikit-Learn libraries. (We will make the code available.)

[4] archive.ics.uci.edu/ml/
[5] https://www.csie.ntu.edu.tw/~cjlin/libsvmtools/datasets/

TABLE I
SUMMARY STATISTICS OF DATASETS USED IN EXPERIMENTS. FOR EACH DATASET WE SHOW THE TRAIN/TEST SAMPLE SIZE, FEATURE SIZE, AND THE RATIO OF NEGATIVE SAMPLES TO POSITIVE SAMPLES IN THE TRAINING SET.

| Dataset | # Samp. | # Feat. | $T_- / T_+$ |
|---|---|---|---|
| a1a | 1.6K / 30.9K | 123 | 3.06 |
| a9a | 32.5K / 16.2K | 123 | 3.15 |
| amazon | 750 / 750 | 10,000 | 2.33 |
| bank | 20.6K / 20.6K | 100 | 7.88 |
| codrna | 29.8K / 29.8K | 8 | 2.00 |
| german | 500 / 500 | 24 | 2.33 |
| ijcnn | 50K / 92K | 22 | 9.30 |
| madelon | 2,000 / 600 | 500 | 1.00 |
| mnist | 60K / 10K | 780 | 2.30 |
| mushrooms | 4K / 4K | 112 | 0.93 |
| phishing | 5.5K / 5.5K | 68 | 0.79 |
| svmguide3 | 642 / 642 | 21 | 2.80 |
| usps | 7.2K / 2K | 256 | 2.61 |
| w1a | 2.5K / 47.2K | 300 | 33.40 |
| w7a | 25K / 25K | 300 | 32.40 |
| avazu app | 12.6M / 2M | 10,000 | 8.33 |
| avazu site | 23.6M / 2.6M | 10,000 | 4.06 |
| criteo | 45.8M / 6M | 10,000 | 2.92 |

### A. Simulation Study

For the simulations we consider the binary hypothesis testing problem of Eq. (1), which can represent, for example, a radar or communication channel setting. In particular, we employ Gaussian mixtures as data generating distributions. Namely, for hypothesis-$i$ a $K$-component Gaussian mixture is given by

$$p_i(\boldsymbol{x}) = \sum_{k=1}^{K} c_{ik} \, (2\pi)^{-d/2} \, |\boldsymbol{\Sigma}_{ik}|^{-1/2} \times \exp\left\{-\frac{1}{2}(\boldsymbol{x} - \boldsymbol{\mu}_{ik})^\top \boldsymbol{\Sigma}_{ik}^{-1}(\boldsymbol{x} - \boldsymbol{\mu}_{ik})\right\}, \quad (36)$$

where the weights $c_{ik}$ are convex combination coefficients to ensure the function is a valid pdf. This distribution is completely characterized by the weights, means, and covariances. For hypothesis-$i$ let $c_i$, $\boldsymbol{\mu}_i$ and $\boldsymbol{\Sigma}_i$ denote the *set* of these parameters. For our experiments $k \in \{1, 2, 3\}$ and:

- $k = 1$: We set $c_0 = \{1\}$, $\boldsymbol{\mu}_0 = \{-\mathbf{0.1}\}$, $\boldsymbol{\Sigma}_0 = \{\boldsymbol{I}\}$ and $c_1 = \{1\}$, $\boldsymbol{\mu}_1 = \{\mathbf{0.1}\}$, $\boldsymbol{\Sigma}_1 = \{\boldsymbol{I}\}$.
- $k = 2$: We set $c_0 = \{0.9, 0.1\}$, $\boldsymbol{\mu}_0 = \{-\mathbf{0.1}, \mathbf{0.1}\}$, $\boldsymbol{\Sigma}_0 = \{\boldsymbol{I}, \boldsymbol{I}\}$ and $c_1 = \{0.1, 0.9\}$, $\boldsymbol{\mu}_1 = \{-\mathbf{0.1}, \mathbf{0.1}\}$, $\boldsymbol{\Sigma}_1 = \{\boldsymbol{I}, \boldsymbol{I}\}$.
- $k = 3$: We set $c_0 = \{0.8, 0.1, 0.1\}$, $\boldsymbol{\mu}_0 = \{-\mathbf{0.1}, \mathbf{0}, \mathbf{0.1}\}$, $\boldsymbol{\Sigma}_0 = \{\boldsymbol{I}, \boldsymbol{I}, \boldsymbol{I}\}$ and $c_1 = \{0.1, 0.1, 0.8\}$, $\boldsymbol{\mu}_1 = \{-\mathbf{0.1}, \mathbf{0}, \mathbf{0.1}\}$, $\boldsymbol{\Sigma}_1 = \{\boldsymbol{I}, \boldsymbol{I}, \boldsymbol{I}\}$

TABLE II
COMPARISONS OF ALGORITHMS ON SIMULATED DATA THE PERFORMANCE OF MBA-$\ell_2$, ONLR, AND ADAAUC ARE REPORTED FOR $k \in \{1, 2, 3\}$ AND SR $\in \{1\%, 10\%, 100\%\}$. THE SYMBOLS FILLED/EMPTY CIRCLE INDICATE THAT MBA IS (STATISTICALLY) SIGNIFICANTLY BETTER/WORSE.

| Distribution | SR | Neyman-Pearson | MBA-$\ell_2$ | ONLR | AdaAUC |
|---|---|---|---|---|---|
| 1-Component Gaussian Mixture | 1 % | | 87.43 ± 0.69 | • 86.79 ± 0.99 | • 80.94 ± 2.77 |
| | 10 % | 92.13 | 91.44 ± 0.10 | • 90.54 ± 0.29 | • 90.25 ± 0.31 |
| | 100 % | | 91.88 ± 0.04 | • 90.69 ± 0.24 | • 91.70 ± 0.07 |
| 2-Component Gaussian Mixture | 1 % | | 80.15 ± 0.68 | • 77.64 ± 1.41 | • 69.75 ± 3.24 |
| | 10 % | 83.71 | 83.15 ± 0.10 | • 80.19 ± 0.69 | • 80.12 ± 0.69 |
| | 100 % | | 83.47 ± 0.04 | • 80.19 ± 0.69 | • 83.01 ± 0.11 |
| 3-Component Gaussian Mixture | 1 % | | 76.39 ± 0.47 | • 73.07 ± 1.06 | • 66.38 ± 1.95 |
| | 10 % | 80.22 | 79.52 ± 0.11 | • 75.94 ± 0.64 | • 76.72 ± 0.33 |
| | 100 % | | 79.93 ± 0.11 | • 75.91 ± 0.83 | • 79.61 ± 0.06 |

As it can be seen the distributions we choose for the two hypotheses are symmetric across the origin. As the number of components increase the distributions get less interspersed, making separation more challenging. All covariances are set to be unit-variance and isotropic; we finally note that the bold numbers for the mean sets correspond to a vector of the bold entry replicated. For our experiments, for each value of $k$ we form 50 training sets, where for each dataset we sample 20,000 points from the generative distribution. Furthermore we create an imbalanced dataset, where roughly 90% of the data has label 0 (i.e. sampled from hypothesis 0). We also create a separate test set, with 100,000 samples and the same imbalance ratio.

Since the generative distributions are assumed known in this setting, we can analytically derive the Neyman Pearson detector which maximizes the AUC. For any given $k$, the NP detector computes the scores through $p_1(\boldsymbol{x})/p_0(\boldsymbol{x})$. Note that the NP-detector does not require any training data, as the generating distributions are already known. A particularly interesting case is $k = 1$. Here the log likelihood is [6]

$$\log \frac{p_1(\boldsymbol{x})}{p_0(\boldsymbol{x})} = \boldsymbol{x}^\top (\boldsymbol{\mu}_1 - \boldsymbol{\mu}_0) + \frac{1}{2}(\boldsymbol{\mu}_0^\top \boldsymbol{\mu}_0 - \boldsymbol{\mu}_1^\top \boldsymbol{\mu}_1) \quad (37)$$

Since the constant term does not affect AUC, we see that the optimum ranking rule is a linear function of $\boldsymbol{x}$. So for this specific case, learning with a linear discriminant (i.e. $f(\boldsymbol{x}) = \boldsymbol{w}^\top \boldsymbol{x}$) is consistent with the generating class [16]. When $k > 1$ the NP rule is no longer linear; and positing a linear discriminant we make a high-SNR assumption.

---
[6] In fact the linearity holds for arbitrary $\Sigma$ as long as it is shared by both hypotheses.

For the comparisons in this section we use MBA-$\ell_2$, ONLR, and AdaAUC, as the latter two typically have the best competitive performance against the former. We run the experiments for three different sample ratio (SR), namely SR $\in \{1\%, 10\%, 100\%\}$; this represents the percentage of available data points used for training. As mentioned above we average the generalization performance over 50 training samples, and report the average values and standard deviations. Table II shows the resulting AUC values and Figure 2 displays the corresponding ROC curves.

Firstly note that, as $k$ increases, the AUC value achieved by the optimal NP rule decreases; this shows that adding more mixture components progressively make the problem harder. As we increase the SR, all three learning algorithms improve, as expected. However, we can see that the starting point for MBA is significantly higher than the the other two. In particular, AdaAUC is worse for small sample sizes. This shows the difficulty of optimizing a bivariate loss function as opposed to the univariate logistic loss of ONLR. The particular difficulty comes from selecting the step size, and for smaller number of samples the stochastic gradient cannot efficiently optimize. Being learning-rate and gradient free, MBA does not suffer from these drawbacks and always performs better than ONLR.

We also use pairwise $t$-test to assess the statistical significance of results, using 95% confidence level, as proposed and used by [8] initially for this problem. For all cases considered we see that MBA achieves better results than its competitors with significance. This is true even when $SR = 100\%$ and the average values are close, as the standard deviations are low and a high number of experiments performed.

Interestingly, comparing the performance of NP and MBA we see that in all three cases the achieved AUC is quite close. This is the case even when $k > 1$; therefore even if the optimal scores are a nonlinear function of $\boldsymbol{x}$ for these cases, the linear approximation is still quite appropriate. As the SNR decreases this approximation will be less realistic; however even the optimal detector can perform quite poor in that regime, so for practical purposes we argue that linearity is a reasonable assumption.

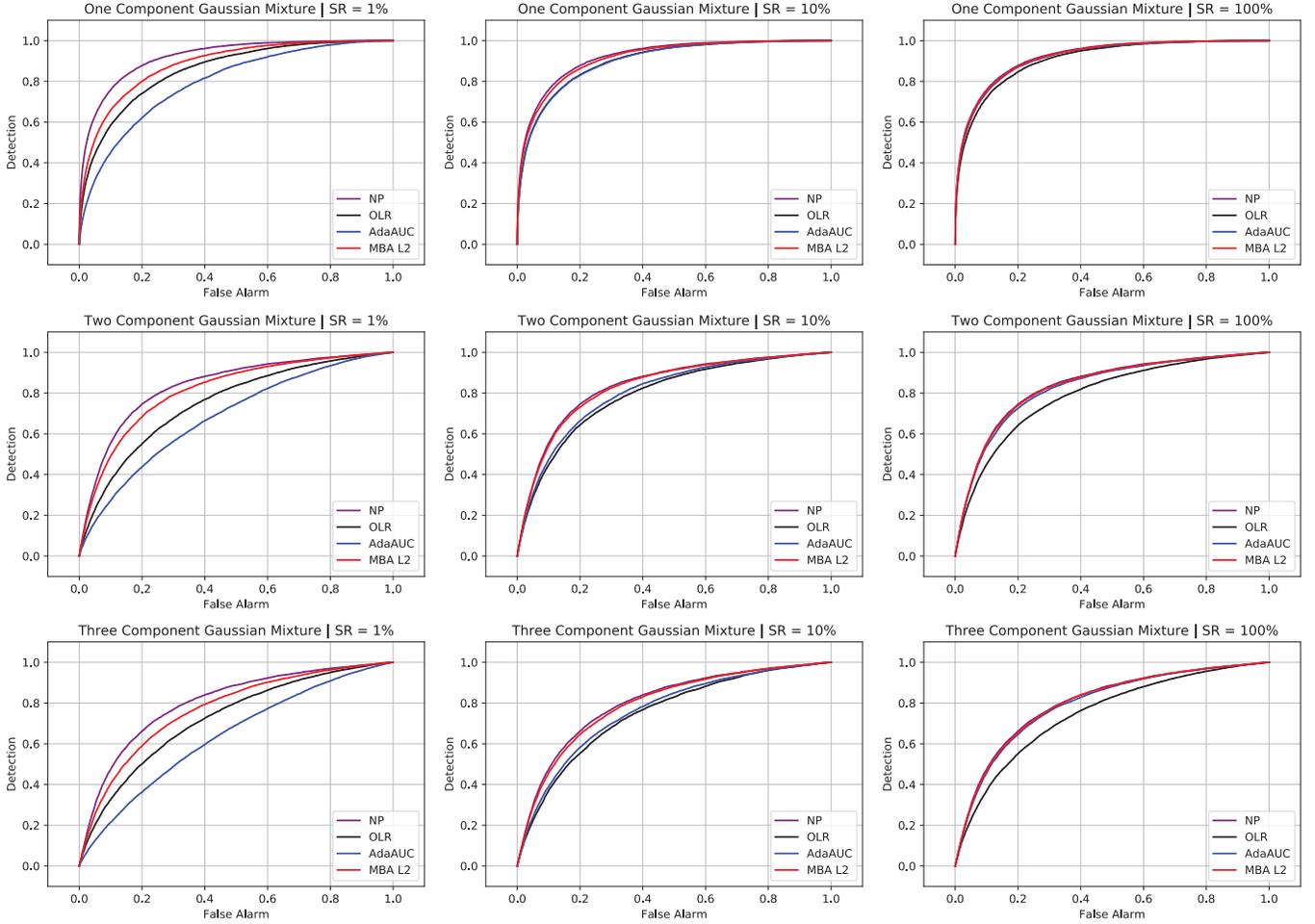

Fig. 2. The ROC curves obtained by the Neyman-Pearson detector and three learning algorithms on the simulated data. The rows are in increasing order of mixture components ($k$) and the columns are in increasing order of sample ratio (SR).

## B. UCI and LIBSVM Benchmark Data

In this section we experiment with 15 frequently used benchmark datasets from the UCI and LIBSVM repositories which we summarize in Table I. It can be seen that the chosen datasets cover a wide range of sample/feature sizes. The distribution of samples vary from being linearly separable to highly nonlinear. The datasets also exhibit significant differences in label imbalance. In terms of the features present, datasets fall into three categories: numerical only, categorical only, and mixed. We use the train/test splits provided in LIBSVM website; if splits are not available we use 50/50 splitting with stratification. For multiclass datasets we map the classes to binary labels.

Table III shows the AUC values obtained by six competing algorithms on benchmark datasets. Here the results are reported along with standard deviations. In addition, we once again conduct a pairwise t-test with 95% significance level. To perform this test, we compare each algorithm in the last four columns to the two MBA algorithms in the first two columns. If MBA performs significantly better/worse we represent this with a filled/empty circle. Table III shows that there is a clear benefit in using the proposed MBA, whereas the recent AdaAUC is the second best competitor. The mini-batch processing phase of MBA is learning-rate free and this brings an important advantage. AdaAUC adapts the gradient steps, while we used the learning rate $\mathcal{O}(1/\sqrt{t})$ for all other stochastic gradient algorithms, which performed well with this choice. However, though MBA does not need this parameter, it still performs significantly better than AdaAUC in 9/15 cases. It is also worth noting that MBA-$\ell_1$ obtains 100% AUC for the mushrooms data, and for the svmguide3 dataset MBA is at least 9% better than the others. While logistic regression does not directly optimize AUC, it is frequently used in practice, where AUC is the main metric, as it typically has competitive performance. Here we see that logistic regression has a decent performance as well, and in fact beats MBA on the a9a data.

Another important performance measure is the ability to rank as a function of sample size. We show comparisons for this in Figure 3. We see that the stochastic gradient based methods keep improving as the sample size increases, whereas MBA has a relatively steady performance, and it converges faster. This indicates that for these benchmark datasets MBA can already construct a good approximation of the global problem at this point. This result is not surprising given Theorem 4, as good performance is independent of the number of pairs or instances, and only related to the dimensionality of the optimization problem to be solved. In the first panel of Figure 3 the best performer is logistic regression although the difference is rather small. In the second plot, MBA-$\ell_2$ gives the best result, although AdaAUC is good as well. For the other two plots, both MBA methods have a clear advantage beginning to end.

TABLE III
COMPARISON OF ALGORITHMS ON 15 BENCHMARK DATASETS FROM UCI AND LIBSVM REPOSITORIES. THE SYMBOLS FILLED/EMPTY CIRCLE INDICATE ONE OF THE MBA IS (STATISTICALLY) SIGNIFICANTLY BETTER/WORSE.

| Dataset | MBA-$\ell_2$ | MBA-$\ell_1$ | OLR | SOLR | OAM | AdaAUC |
|---|---|---|---|---|---|---|
| a1a | 88.98 ± 0.14 | 88.67 ± 0.15 | • 88.64 ± 0.27 | • 88.04 ± 0.14 | • 87.61 ± 0.45 | • 88.51 ± 0.34 |
| a9a | 89.97 ± 0.01 | 89.97 ± 0.02 | ∘ 90.17 ± 0.03 | • 89.88 ± 0.03 | • 89.30 ± 0.22 | 89.99 ± 0.04 |
| amazon | 77.12 ± 0.44 | 71.35 ± 2.50 | • 69.90 ± 2.21 | • 71.87 ± 0.74 | • 60.23 ± 3.90 | • 74.97 ± 0.89 |
| bank | 93.22 ± 0.06 | 93.22 ± 0.03 | • 82.89 ± 0.21 | • 80.23 ± 0.33 | • 81.51 ± 0.49 | • 89.46 ± 0.12 |
| codrna | 97.68 ± 0.00 | 97.63 ± 0.01 | • 95.69 ± 0.19 | • 92.36 ± 0.85 | • 97.31 ± 0.12 | • 94.34 ± 0.40 |
| german | 80.34 ± 0.80 | 80.41 ± 0.64 | • 76.39 ± 1.69 | • 75.07 ± 1.22 | • 74.59 ± 1.79 | • 77.83 ± 1.29 |
| ijcnn | 90.53 ± 0.05 | 90.40 ± 0.07 | • 89.50 ± 0.53 | • 88.93 ± 0.48 | • 88.52 ± 1.76 | 90.59 ± 0.28 |
| madelon | 62.39 ± 0.44 | 62.34 ± 0.51 | 61.97 ± 0.69 | • 61.81 ± 0.48 | • 60.64 ± 0.44 | 61.82 ± 1.58 |
| mnist | 95.81 ± 0.02 | 95.77 ± 0.02 | • 95.63 ± 0.27 | • 95.49 ± 0.12 | • 94.82 ± 0.23 | • 95.47 ± 0.09 |
| mushrooms | 100.00 ± 0.00 | 100.00 ± 0.00 | 99.88 ± 0.03 | • 99.73 ± 0.07 | • 99.62 ± 0.28 | • 99.98 ± 0.00 |
| phishing | 98.32 ± 0.01 | 98.32 ± 0.05 | 98.49 ± 0.01 | 98.38 ± 0.03 | • 98.08 ± 0.27 | 98.36 ± 0.02 |
| svmguide3 | 81.16 ± 0.80 | 82.05 ± 0.66 | • 63.80 ± 0.81 | • 57.65 ± 2.98 | • 66.97 ± 3.45 | • 69.14 ± 1.95 |
| usps | 95.89 ± 0.04 | 95.83 ± 0.06 | • 95.82 ± 0.15 | • 95.71 ± 0.07 | • 94.65 ± 0.53 | • 95.74 ± 0.13 |
| w1a | 92.28 ± 0.23 | 91.21 ± 0.36 | • 84.81 ± 1.34 | • 79.84 ± 1.15 | • 87.83 ± 1.59 | • 90.70 ± 0.67 |
| w7a | 96.27 ± 0.07 | 96.17 ± 0.08 | • 93.05 ± 0.29 | • 89.27 ± 0.82 | • 93.92 ± 0.55 | • 95.09 ± 0.26 |
| Win/Tie/Loss | - | - | 11/3/1 | 14/1/0 | 15/0/0 | 11/4/0 |

### C. Large-scale Web Click Data

For this last part of experiments we use large scale datasets where the task is click through rate (CTR) prediction. Estimating user clicks in web advertising is one of the premier areas of AUC optimization. As the number of users who click a given ad is typically low, the task naturally manifests itself as distinguishing click from non-click. The datasets used come from Avazu and Criteo, available at LIBSVM; we once again summarize them at the end of Table I. It is worthwhile to note that these datasets are an order of magnitude larger than the ones used in previous studies, showing the scaling benefit MBA brings. This time we shuffle and split the entire dataset into chunks of 100 (Avazu App) and 200 (Avazu Site and Criteo). We then make a single pass over these chunks with randomized sampling and report the results. For these datasets the variation across different runs is very small, as the inputs are very uniform. Therefore we do not show the confidence intervals in the bar charts, but note that all results are statistically significant. For the CTR problem, a 0.1% improvement in AUC is considered significant, whereas an increase of 0.5% results in noticable revenue gain.

In Figure 4 we show the AUC performance of seven algorithms. For the Avazu App data, MBA-$\ell_2$ gives the best results while for Avazu Site and Criteo all MBA algorithms give similar results. Comparing the proposed MBA with the best non-MBA algorithm, the performance improvements are 1.20%, 0.43% and 0.54%. The mini-batch gradient descent algorithms do not perform as well, especially when the regularization parameter is small, and they get better as this parameter increases. For these experiments the step size is $\mathcal{O}(1/\sqrt{t})$, and while logistic regression has good performance with this choice, optimizing pairwise losses seems less robust. As the regularization increases the variation in the gradients decreases, which helps improve the AUC scores. We also experiment with a small constant step size, and this yields similar results. On the other hand MBA does not require this parameter at all. For this reason our algorithm is quite suitable for large-scale problems.

Another important concern is the running time. Here, we do not make a relative comparison, instead we state how much time it takes to find the result. This is because, comparing the running time to logistic regression is not very informative; if a sequential logistic regression is implemented in Python script, then the mini-batch algorithm is roughly 10 times faster, as sequential processing is slow. However, if an optimized package is used, then it can be 100 times faster than MBA, as the underlying code is optimized. For this reason we show the running time of the vanilla implementation of MBA in Figure 5. As it can be seen, even for the Criteo dataset, which contains the largest number of instances, the runtime is under an hour. As we briefly mentioned in Section III, the mini-batch portion of MBA can be distributed without loss of accuracy, therefore using cluster computing, MBA can easily scale to billion-sample datasets, which are several orders of magnitude larger than the datasets that can be handled by sequential methods.

## VI. CONCLUSION

This paper has introduced a fast algorithm to optimize the AUC metric. Our proposed approach, called MBA, uses the specific structure of the squared pairwise surrogate loss function. In particular, it is shown that one can approximate the global risk minimization problem simply by approximating the first and second moments of pairwise differences of positive and negative inputs. This suggests an efficient mini-batch scheme, where the moments are estimated by U-statistics. MBA comes with theoretical guarantees, and importantly the number of samples required for good performance is independent of the number of pairs present, which is typically a very large number. Our experiments demonstrate the advantages of MBA in terms of speed and performance. We think MBA would be particularly useful for applications where AUC is the prime metric, and the data size is massive and parallel processing is necessary.

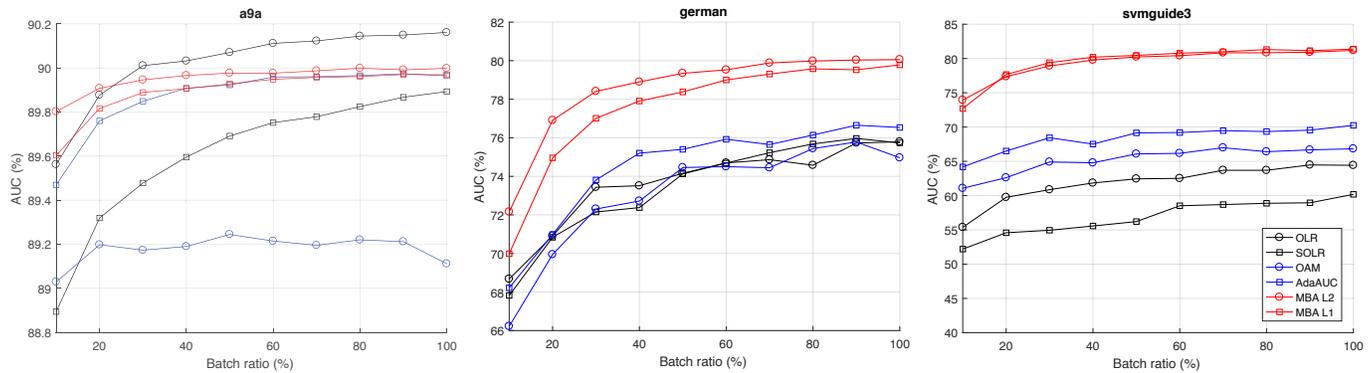

Fig. 3. AUC performance of six algorithms as a function of sample size for a9a, german, and svmguide3 selected from LIBSVM.

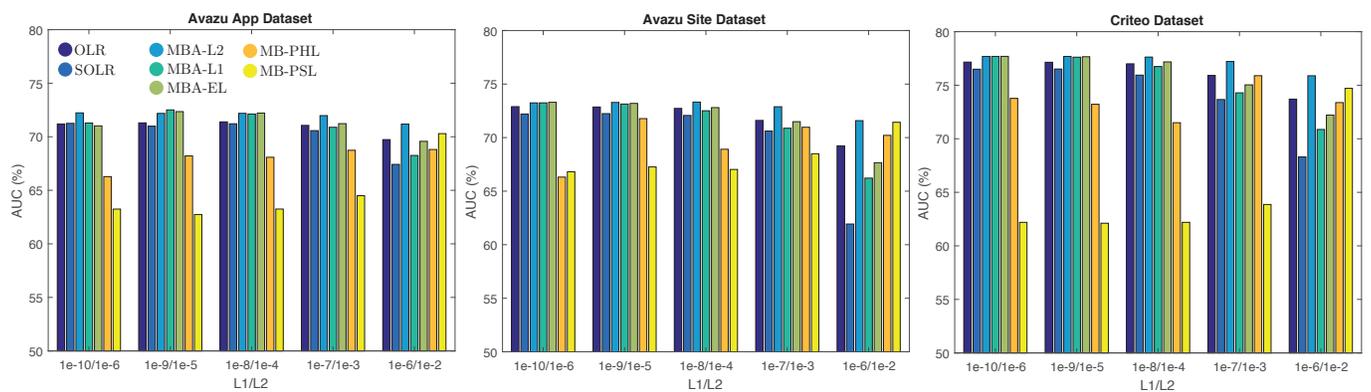

Fig. 4. AUC achieved by all algorithms on the Avazu App, Avazu Site, and Criteo datasets. Here the performance is plotted as a function of regularization parameters. The elastic net uses one half of $\ell_1$-penalty for both $\ell_1$ and $\ell_2$ regularization.

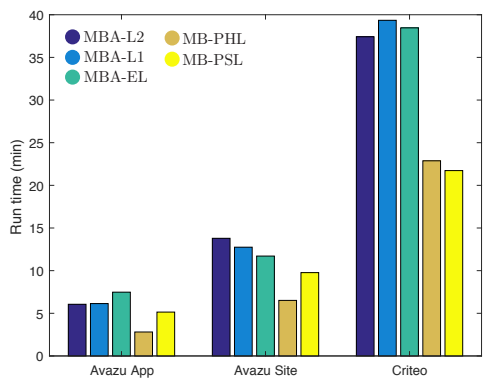

Fig. 5. Runtime comparison of MBA with MB-PSL and MB-PHL. As the latter two only require a gradient computation they are faster than MBA, but with significantly reduced performance. On the other hand MBA can process ten million samples under an hour, which shows the scalability of this approach.